\documentclass[10pt,journal,compsoc]{IEEEtran}

\usepackage{stfloats}
\usepackage{amsfonts}
\usepackage{amssymb}
\usepackage{algorithm}
\usepackage{algpseudocode}
\usepackage{graphicx}
\usepackage{subfigure}
\usepackage{color}
\usepackage{float}
\usepackage{longtable}
\usepackage{amsmath}
\usepackage{multirow}

\ifCLASSOPTIONcompsoc
  \usepackage[nocompress]{cite}
\else
  \usepackage{cite}
\fi

\begin{document}

\title{Multi-modal Face Pose Estimation with Multi-task Manifold Deep Learning}

\author{Chaoqun~Hong,
        Jun~Yu,~\IEEEmembership{Member,~IEEE,}
        and~Jian~Zhang

\IEEEcompsocitemizethanks{\IEEEcompsocthanksitem J. Yu is with the School of Computer Science, Hangzhou Dianzi University, Hangzhou, 310018, China.\protect\\
E-mail: yujun@hdu.edu.cn
\IEEEcompsocthanksitem C. Hong is with the College of Computer and Information Engineering, Xiamen University of Technology, Xiamen, 361024, China.}
\thanks{Manuscript received xx, xx.}}

\markboth{IEEE Transactions on Knowledge and Data Engineering,~Vol.~xx, No.~xx, xx~xx}%
{Shell \MakeLowercase{\textit{et al.}}: Bare Demo of IEEEtran.cls for Computer Society Journals}

\IEEEtitleabstractindextext{%
\begin{abstract}
Human face pose estimation aims at estimating the gazing direction or head postures with 2D images. It gives some very important information such as communicative gestures, saliency detection and so on, which attracts plenty of attention recently. However, it is challenging because of complex background, various orientations and face appearance visibility. Therefore, a descriptive representation of face images and mapping it to poses are critical. In this paper, we make use of multi-modal data and propose a novel face pose estimation method that uses a novel deep learning framework named Multi-task Manifold Deep Learning ($M^2DL$). It is based on feature extraction with improved deep neural networks and multi-modal mapping relationship with multi-task learning. In the proposed deep learning based framework, Manifold Regularized Convolutional Layers (MRCL) improve traditional convolutional layers by learning the relationship among outputs of neurons. Besides, in the proposed mapping relationship learning method, different modals of face representations are naturally combined to learn the mapping function from face images to poses. In this way, the computed mapping model with multiple tasks is improved. Experimental results on three challenging benchmark datasets DPOSE, HPID and BKHPD demonstrate the outstanding performance of $M^2DL$.
\end{abstract}

\begin{IEEEkeywords}
human face pose estimation, multi-task learning, deep learning, manifold learning.
\end{IEEEkeywords}}

\maketitle

\IEEEdisplaynontitleabstractindextext

%
\IEEEpeerreviewmaketitle

\IEEEraisesectionheading{\section{Introduction}\label{sec:introduction}}
\IEEEPARstart{E}{stimating} the pose of human faces aims at achieving the gazing direction or head postures with images. Recently, it attracts plenty of attention since it gives some very important information such as communicative gestures, saliency detection and so on. Therefore, it is critical in human activity analysis, human-computer interface and some other applications. Generally speaking, according to the routine of estimation by mapping images to poses, current methods focus on feature representation of head images and mapping the representation to head poses \cite{siriteerakul2010head} \cite{benabdelkader2010robust}.

Feature representation is the key to the problem of face pose estimation and feature mapping often depends on it. Therefore, a descriptive representation is critical. Recently, some researchers propose to represent head images in different feature spaces that have more discriminatory property for face pose independent of people. Chen and Odobez \cite{chen2012we} proposed the state-of-the-art method for unconstrained coupled head-pose and body-pose estimation in surveillance videos. They used multi-level Histogram of Oriented Gradients (HOG) \cite{dalal2005histograms} for the head and body pose features and extracted a feature vector for an adaptive classification using high dimensional kernel space methods. A similar idea is applied by Flohr et al. to jointly estimate head poses and body poses \cite{Flohr2015A}. These techniques are quite general and do not depend on the heads being in near frontal poses unlike current Human-Computer Interface (HCI) techniques. Nevertheless the high degree of error or uncertainties that arise from these methods, render them unsuitable for the tasks like fine grained human interaction or attention modeling.

To obtain descriptive representation, deep learning architectures \cite{srivastava2014dropout}\cite{shao2014learning} have been efficient in exploring hidden representations in natural images and have achieved proven success in a variety of vision applications. Therefore, we look into deep learning. It is naturally data-driven and powerful in concept abstraction. In this way, we assume that it is able to describe the face images and connect them with the corresponding poses with sufficient data. For example, an autoencoder \cite{bengio2009learning} is an efficient unsupervised feature-learning method in which the internal layer acts as a generic extractor of inner image representations. A double-layer structure, which efficiently maps the input data onto appropriate outputs, is obtained by using a multilayer perceptron. In addition, deep learning can exploit parallel GPU computation and deliver high speeds in the forward pass. These advantages make deep models an attractive option for handling the face pose estimation problem. However, the neurons in neural networks are often considered separately, which limits the performance.

Even with descriptive representation human face pose estimation is still difficult since human faces are usually captured at very low resolution and appear blurred. In addition, the rotation of human faces is complicated and the facial features are usually invisible in most of the scenarios. Therefore, employing a single camera view or a single type of input is often insufficient for studying peoples behavior in large environments. A handful of approaches \cite{munoz2012multi}\cite{zabulis20093d}\cite{voit2009system} have exploited multi-view images to achieve robust pose estimation. Yet, most of them estimate face pose of a person rotating in place. Rajagopal et al. proposed a method to estimate peoples 3D head orientation as they freely move around in naturalistic settings such as parties, museums and supermarkets \cite{rajagopal2014exploring}. Inspired by multi-view ideas, Mukherjee and Robertson proposed a multimodal method \cite{mukherjee2015deep}. It uses both RGB images and depth images to improve the performance. Furthermore, by defining classification in different views as different tasks, Yan et al. proposed a novel framework based on Multi-Task Learning (MTL) for classifying the face pose of a person who moves freely in an environment monitored by multiple, large field-of-view surveillance cameras \cite{yan2016multi}.

In a word, current trend of face pose estimation is using multi-view or multi-modal for learning. In traditional methods, multi-view integration is often over-simplified and the mapping function is assumed to be linear or nonlinear. However, in reality, the correspondence between face images and poses is complicated. It is often ambiguous or even arbitrary. Therefore, we try to improve feature description and mapping by deep learning and multi-task learning respectively. Despite some exciting results with classification or recognition from the related deep learning approaches, the performance can be further improved with the structural information of data considered \cite{Yuan2015Scene}. Manifold learning is widely used to learn the structural information \cite{Raytchev2004Head}. In this way, we develop a big data-driven strategy for face pose estimation in this paper. Specifically, we design a novel deep architecture named Multi-task Manifold Deep Learning ($M^2DL$) for multi-modal human face data. Different from existing face pose estimation methods, the proposed method applies Deep Convolutional Neural Network (DCNN) based feature extraction to represent face images and multi-task learning based modal to construct the mapping relationship from images to poses. The contributions of this paper are summarized below:

\begin{enumerate}
  \item First, we propose a new multi-task learning framework based on Deep Convolution Neural Network (DCNN). In this framework, DCNN-based feature mapping and multi-task learning is connected to obtain a DCNN-Based regression for face pose estimation, which unified the multi-view problem and multi-modal problem in a single model.
  \item Second, in the proposed framework, convolutional layers are improved with manifold regularization, which is called Manifold Regularized Convolutional Layers (MRCL). In the proposed MRCL, the inner relationship of neurons are utilized. In this way, locality properties of neurons can be kept and better feature mapping can be learnt.
  \item Finally, the proposed framework is naturally multi-modal and can be used in different scenarios. We conduct comprehensive experiments to empirically analyze our method on three benchmark datasets. The experimental results validate the effectiveness of our method.
\end{enumerate}

The remainder of this paper is organized as follows. Related works on multi-task learning and manifold learning are reviewed in Section II. Then, the proposed $M^2DL$ is presented in Section III. After that, we demonstrate the effectiveness of $M^2DL$ by experimental comparisons with other state-of-the-art methods in Section IV. We conclude in Section V.

\section{Related Works}
\subsection{Deep learning}
As mentioned before, the traditional routine of face pose estimation with images consists of feature representation and feature mapping.

\begin{enumerate}
  \item Feature representation. In this part, researchers try to represent the images with descriptive features. The pioneering work on face pose estimation was proposed by Robertson and Reid \cite{robertson2006estimating} which used a detector based on template training to classify face poses in eight directional bins. This approach is heavily reliant on skin colour model. Subsequently this template-based technique was extended to a color invariant technique by Benfold et al. \cite{benfold2008colour}. Based on the template features, they proposed a randomized fern classifier for hair face segmentation for matching. This work was later improved by Siriteerakul et al. \cite{siriteerakul2010head} using pair-wise local intensity and colour differences. However, in keeping with all template based techniques in head-pose estimation, these suffer from two major problems: first, it is non-trivial to localize the head in low resolution images; second, different poses of the same person may appear more similar compared to the same head-pose of different persons.
  \item Feature mapping. In this part, researchers try to define the reasonable mapping relationship from images to poses. Previous methods usually employ regression. The regression methods used for face pose are Gaussian process regression (GPR) \cite{marin2014detecting}, support vector regression (SVR) \cite{murphy2007head}, partial least squares (PLS) \cite{sharma2011bypassing} and kernel PLS \cite{al2012partial}. Both \cite{marin2014detecting} and \cite{murphy2007head} estimate the pose angles independently, so several regression functions must be learned, one for each angle, hence correlations between these parameters cannot be taken into account. Another drawback of all kernel methods is that they require the design of a kernel function with its hyper-parameters, which must be either manually selected or properly estimated using non-convex optimization techniques.
\end{enumerate}

Recently, there¡¯s been a great deal of excitement and interest in deep neural networks because they¡¯ve achieved breakthrough results in areas such as computer vision \cite{Krizhevsky2017ImageNet}. In the current big data era, the extensive availability of training images enables deep models to be generic and flexible. In feature representation, deep learning can be used to improve the descriptive ability \cite{Liu20163D} \cite{Demirkus2014}. In feature mapping, non-linear regression approaches like Artificial Neural Networks \cite{gourier2006head}, \cite{stiefelhagen2004estimating} and high-dimensional manifold based approaches \cite{balasubramanian2007biased} \cite{benabdelkader2010robust} try to estimate the face poses in a continuous range.

\subsection{Multi-task learning}

Multi-task learning has recently been employed in image classification \cite{yuan2012visual}, visual tracking \cite{Zhang2013Robust}, multi-view action recognition \cite{Yan2014Multitask} and egocentric daily activity recognition \cite{Yan2015Egocentric}. Given a set of related tasks, MTL \cite{Caruana1997Multitask} seeks to simultaneously learn a set of task-specific classification or regression models. The intuition behind MTL is that a joint learning procedure accounting for task relationships is more efficient than learning each task separately. Traditional MTL methods \cite{Sch2007Multi}\cite{Evgeniou2004Regularized} assume that all the tasks are related and their dependencies can be modeled by a set of latent variables. However, in many real world applications, not all tasks are related, and enforcing erroneous (or non-existent) dependencies may lead to negative knowledge transfer.

Recently, sophisticated methods have been introduced to counter this problem. These methods assume a-priori knowledge (e.g., a graph) defining task dependencies \cite{Chen2011Smoothing}, or learn task relationships in conjunction with task-specific parameters \cite{Kang2011Learning}, \cite{Zhong2012Convex}, \cite{Jalali2010A}, \cite{Zhou2011Clustered}, \cite{Gong2012Robust}. Among these, our work is most similar to \cite{Chen2011Smoothing} and our algorithm adopts two graphs (one defining appearance similarity among grid segments, and the other relating face pose classes) to specify task dependencies.

\subsection{Manifold learning}

Traditionally, images are represented by different features, such as color, texture, shape and so on. Some of these features are high dimensional. High dimensionality results in high computational complexity in pose retrieval, which is called ¡±curse of dimensionality¡±. To overcome it, researchers look into dimension reduction, which is often achieved by manifold learning or subspace learning \cite{Yu2012Local}. It aims at finding the transformation from the original feature space to a low dimensional subspace that retains most of the discriminative information. There are also many existing methods such as Principal Component Analysis (PCA) \cite{Hotelling1933Analysis}, Linear Discriminant Analysis (LDA) \cite{Riffenburgh1960Linear}, Discriminative Locality Alignment (DLA) \cite{Zhang2008Discriminative}, ISOMAP \cite{Tenenbaum2000A}, Locality Sensitive Discriminant Analysis (LSDA) \cite{Cai2007Locality}, Locality Preserving Projections (LPP) \cite{He2003Locality}, Neighborhood Preserving Embedding(NPE) \cite{He2005Neighborhood}, Isometric Projection (IsoP) \cite{Cai2007Isometric} and so on. Raytchev et al. extended the Isomap model to be able to map (high-dimensional) input data points which were not in the training data set into the dimensionality-reduced space found by the model. From this representation, a pose parameter map relating the input face samples to view angles was learnt \cite{Raytchev2004Head}. Aghajanian et al. represented a face with a non-overlapping grid of patches in manifold space. This representation was used in a generative model for automatic estimation of head pose \cite{Aghajanian2009Face}.

The structures of features are important for data mining, especially for multi-view data. However, traditional methods make use of multiple feature by directly concatenating them, which is over-simplified. To solve this problem, researchers also apply manifold learning to combine different types of features \cite{Vadakkepat2008Multimodal}\cite{Luo2012Enriched}. These methods are called multi-view learning methods, which has been comprehensively studied and widely used.

\section{Multi-task Manifold Deep Learning}

\subsection{Overview of the proposed method}

\begin{figure*}[!t]
\centering
\includegraphics[width=0.98\linewidth]{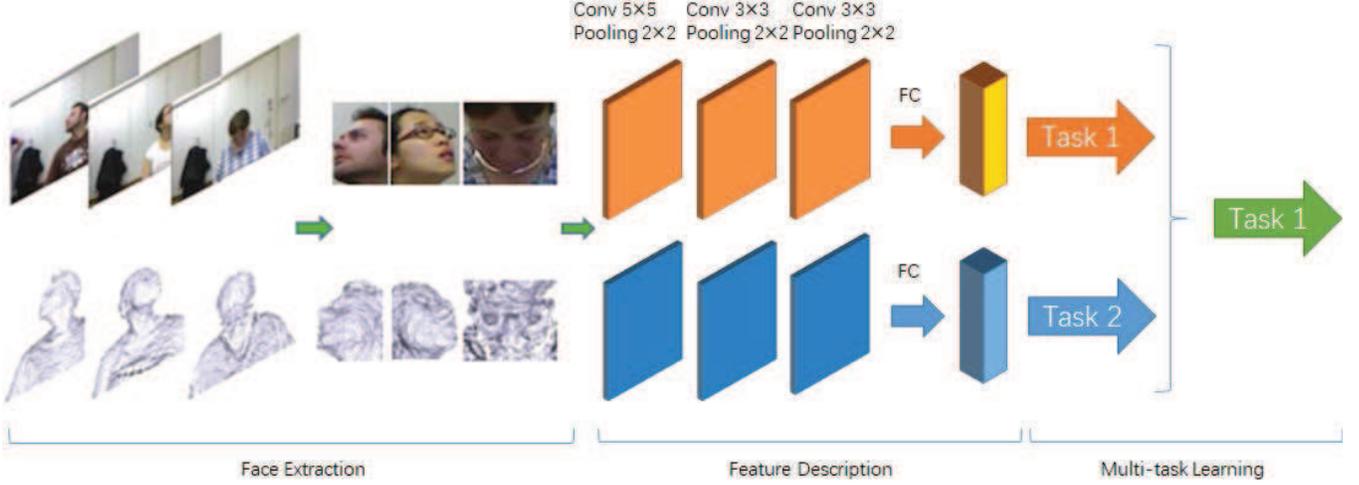}
\caption{The flowchart of the proposed method is shown. Without loss of generality, we use the dataset with color images and depth images as an example. Generally speaking, it consists of three stages. First, we extract clear faces from the images. Second, we use deep neural networks to compute features. For two types of features, we train two networks. In this stage, we propose Manifold Regularized Convolutional Layers. Finally, we connect multi-task learning with neural networks to compute the regression model for face pose estimation. In this stage, the model with each type of features is considered as a task. With this model, we can map image features to estimated poses.}
\label{flowchart}
\end{figure*}

The flowchart of the proposed method is shown in Fig. \ref{flowchart}. To get rid of the influences of background, we should extract faces in images first. This process depends on the definitions of different datasets. In some datasets, the positions and sizes are provided and they can be used directly. However, in some other datasetes, we need face detection or face tracking to get the face area. Then, we use three convolutional layers and one full connected layer to compute the features. Finally, multi-task learning is applied to learn the mapping models from images to poses. In different datasets, the definitions of tasks are also different. For multi-view image datasets, we set a view as a task. For datasets with different modals, we set a modal as a task.

\subsection{The Framework of Multi-task Learning}

As mentioned before, the traditional routine to achieve face pose estimation is mapping images to poses with pre-computed regression models. Therefore, the key is computing a well-defined regression model. In data mining and machine learning, a common paradigm for classification and regression is to minimize the penalized empirical loss:

\begin{equation}
\arg \min_W \ell(W)+\Phi(W),
\end{equation}

\noindent where $W$ is the parameter to be estimated from the training samples, $\ell(W)$ is the loss function and $\Phi(W)$ is the regularization term that encodes task relatedness.

In our application, multi-view face pose estimation with $V$ views can be considered as a multi-task process with $V$ tasks. The training data for $v$-th task can be denoted by $x_i^v,y_i^v$, where $v=1,...,V$, $i = 1,...,N$ and $N$ is the number of samples. $X=x_i^vR^{d1}$ and  $Y=y_i^vR^{d2}$ are image features and face poses respectively, where $d1$ and $d2$ are their dimensions respectively. The goal of multi-task learning can be defined as:

\begin{equation}
\arg \min \sum_{v=1}^V \sum_{i=1}^N \ell(y_i^v,f(x_i^v;w^v))+\Phi(w^v),
\end{equation}

\noindent where $f(x_i^v;w^v)$ is a function of $x_i^v$ and parameterized by a weight vector $w^v$. There are several existing choices of $\ell(\cdot)$. The least square is widely used for regression and the hinge loss is used for classification. $\Phi(w^v)$ is the regularization term that penalizes the complexity of weights. In this way, the objective function can be rewritten as:

\begin{equation}
\arg \min_W \frac{1}{2} \sum_{v=1}^V \parallel Y - \mathcal{F}(X^v;W^v) \parallel ^2+\sum_{v=1}^V \parallel W^v \parallel ^2 ,
\end{equation}

\noindent where $W = w^v$ is the weighted matrix with the same meaning as Eq. (1) and task relatedness is encoded by summing the weights. To solve the above function, the key is how to define an optimized regression function $\mathcal{F}(X^v;W^v)$.

\subsection{Deep Convolution Neural Network based regression}

Deep neural networks has been proven success in image description, especially with multi-task learning \cite{Li2016DeepSaliency}. In our method, we solve $f(\cdot)$ by using the deep neural network. In the deep neural network, this function is called the activation function. In computational networks, the activation function of a node defines the output of that node given an input or set of inputs. In the scenario of the deep neural network, activation functions project $x^v_i$ to higher level representation gradually by learning a sequence of non-linear mappings, which can be defined as:

\begin{equation}
(x^v_i)^0\xrightarrow[W]{\mathcal{R}}(x^v_i)^1\xrightarrow[W]{\mathcal{R}}...\xrightarrow[W]{\mathcal{R}}(x^v_i)^l,
\end{equation}

\noindent where $l$ is the number of layers and $\mathcal{R}$ is the mapping function from input to estimated output.

To optimize the weighted matrix $W$ which contains the mapping parameters, we use a back-propagation strategy. For each echo of this process, the weighted matrix is updated by $\Delta W$, which is defined by:

\begin{equation}
\Delta W = - \eta \frac{\partial \mathcal{Z}}{\partial W}.
\end{equation}

\noindent $\eta$ is the learning rate and $\mathcal{Z}$ is the energy loss of neural networks. Eq. (5) can be further defined as:

\begin{equation}
\frac{\partial \mathcal{Z}}{\partial W} = (y_i^v-\mathcal{R}(x_i^v))(x_i^v)^T.
\end{equation}

\noindent In this way, we try to minimize the differences between the groundtruth $y_i^v$ and the estimated output $\mathcal{R}(x_i^v)$. The back-propagation strategy can be modeled by:

\begin{equation}
(x^v_i)^0\xleftarrow[W]{\mathcal{R}}(x^v_i)^1\xleftarrow[W]{\mathcal{R}}...\xleftarrow[W]{\mathcal{R}}(x^v_i)^l.
\end{equation}

\subsection{Manifold Regularized Convolutional Layers}

Classic convolutional neuron networks consist of alternatively stacked convolutional layers and spatial pooling layers. The convolutional layers generate feature maps by linear convolutional filters followed by nonlinear activation functions (rectifier, sigmoid, tanh, etc.). Rectified Linear Units (ReLU) is defined by:

\begin{equation}
f_r(x)=
\left\{
\begin{array}{l}
0 \mbox{ , }x<0 \\
x \mbox{ , }x \ge 0
\end{array}.
\right.
\end{equation}

Sigmoid is defined by:

\begin{equation}
f_s(x) = \frac{1}{1+e^{-x}}
\end{equation}

Tanh is defined by:

\begin{equation}
f_t(x) = \frac{2}{1+e^{-2x}}-1
\end{equation}

The above activation functions are non-linear and widely used. Taking the Rectified Linear Units (ReLU) as an example, the feature map can be calculated as follows:

\begin{equation}
F(X^v;W^v)=\mathcal{R}(x_i^v)=max(0,w^vx_i^v)=
\left\{
\begin{array}{l}
0 \mbox{ , }x_i^v<0 \\
w_k^Tx_i^v \mbox{ , }x_x^v \ge 0
\end{array}.
\right.
\end{equation}

\begin{figure*}[!t]
\centering
\includegraphics[width=0.98\linewidth]{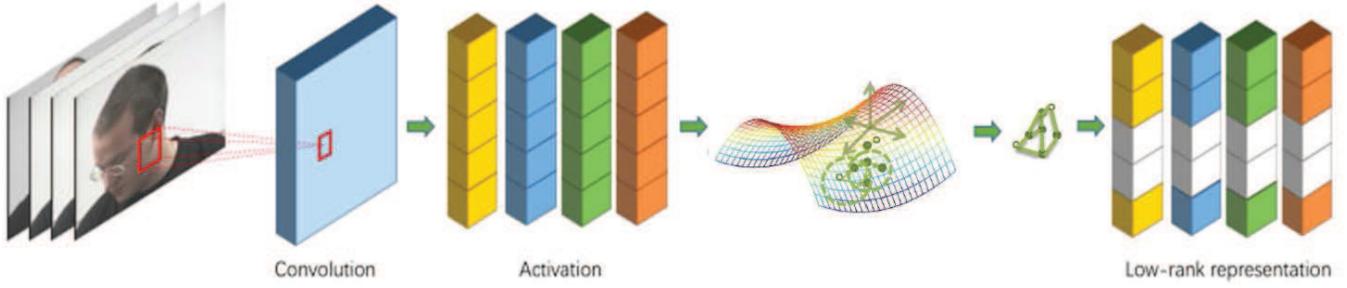}
\caption{The process of Manifold Regularized Convolutional Layer is shown. After convolution and activation of traditional CNNs, we add manifold regularization with low-rank representation.}
\label{cnn}
\end{figure*}

Traditional CNN implicitly makes the assumption that the latent concepts are linearly separable. However, the data for the same concept often live on a nonlinear manifold, therefore the representations that capture these concepts are generally highly nonlinear function of the input. In the proposed Manifold Regularized Convolutional Layers (MRCL), we add manifold regularization to impose the locality constraints, which is shown in Fig. \ref{cnn}.

In MRCL, the output is defined as:

\begin{equation}
F(X^v;W^v)=M(x^v),
\end{equation}

\noindent where $x^v$ represents all the samples in Task $v$ and $M(x^v)$ is the output of $R(x^v)$ with manifold structure. In this way, Eq. (3) can be re-rewritten as:

\begin{equation}
\arg \min_W \frac{1}{2}\sum_{v=1}^V \parallel Y-M(x^v) \parallel ^2+\sum_{v=1}^V \parallel W^v \parallel _2^2,
\end{equation}

To learn the manifold structure of $\mathcal{R}(x^v)$ and compute $M{\cdot}$, many existing methods can be used, such as subspace learning and manifold learning. Recently, low-rank sparse learning attracts plenty of attention. In order to recover the low-rank matrix $X_0$ from the given observation matrix $X$ corrupted by errors $E_0(X_0+E_0)$, it is straightforward to consider the following regularized rank minimization problem:

\begin{equation}
\min_{D,E} rank(D) + \lambda \parallel E \parallel _l, \mbox{ s. t. } X = D+E,
\end{equation}

\noindent where $\lambda > 0$ is a parameter and $\parallel \cdot \parallel_l$ indicates certain regularization strategy, such as Frobenius Form, the $l_0$ norm and the $l_{2,0}$ norm. To better handle the mixed data, Liu et. al suggested a more general rank minimization problem defined as follows:

\begin{equation}
\min_{Z,E} rank(Z) + \lambda \parallel E \parallel _l, \mbox{ s. t. } X = AZ+E,
\end{equation}

\noindent where $A$ is a dictionary that linearly spans the data space and the minimized $Z^*$ (with regard to the variable $Z$) is the lowest rank representation of data $X$ with respect to a dictionary $A$. To solve (12), we use the $l_{2,1}$ norm since it ensures the row sparsity. In this way, a low-rank recovery to $X_0$ is obtained by solving the following convex optimization problem:

\begin{equation}
\min_{Z,E} rank(Z) + \lambda \parallel E \parallel_{2,1}, \mbox{ s. t. } X = AZ+E,
\end{equation}

To solve it, the Augmented Lagrange Multiplier method can be used. In this way, it can be converted to the following equivalent problem:

\begin{equation}
\min_{Z,E,J} rank(Z) + \lambda \parallel E \parallel_{2,1}, \mbox{ s. t. } X = AZ+E,Z=J.
\end{equation}

It can be solved by the ALM method, which minimizes the following augmented Lagranrian function:

\begin{equation}
L=\parallel E \parallel _*+\lambda \parallel E \parallel_{2,1}+tr(Y_1^T(X-AZ-E))+tr(Y_2^T(Z-J))+\frac{\mu}{2}(\parallel X-AZ-E) \parallel_F^2+\parallel Z-J \parallel^2_F.
\end{equation}

The above problem is unconstrained. So, it can be minimized with respect to $J$, $Z$ and $E$, respectively, by fixing the other variables and then updating the Lagrange multipliers $Y1$ and $Y2$, where $\mu>0$ is a penalty parameter.

\subsection{Implementation Details}

In the implementation of Multi-task Deep Learning with Manifold Regularized Convolutional Layers, we use three convolutional layers and each is followed by max pooling. Finally, a fully connected layer that has 512 neurons is used to get feature mapping. In the multi-task learning stage, we optimize (3) with the least trace method. With this key structure of deep convolutional network, the details of algorithms are shown in Algorithm 1. To implement the proposed deep neural network, Caffe is used and works on a workstation with 4 Titan X (Pascal) GPUs.

\begin{algorithm}[htb]
\caption{Details of Multi-task Deep Learning with Manifold Regularized Convolutional Layers}
\label{mmdl}
\begin{algorithmic}[1]
\Require Multi-view Head Images $X = x^v \in R^{d1}$
\Ensure Face Poses represented by pan, tilt and so on. $Y = y^v \in R^{d2}$
\State Stage 1:
\ForAll {each view}
\State The first convolutional layer, with $32$ kernels of size $5 \times 5 \times 1$, followed by $2 \times 2$ max pooling;
\State The second convolutional layer, with $32$ kernels of size $3 \times 3 \times 32$, followed by $2 \times 2$ max pooling;
\State The third convolutional layer, with $24$ kernels of size $3 \times 3 \times 32$, followed by $2 \times 2$ max pooling;
\State The fully connected layer that has $512$ neurons, $M(x^v)$;
\EndFor
\State Stage 2:
\State Solving Eq. (3) with a specific multi-task loss method;
\State Mapping testing images to poses with optimized parameter $W$;
\\
\Return mapped face poses;
\end{algorithmic}
\end{algorithm}

\section{Experimental Results and Discussions}

\subsection{Benchmark datasets}

In this section, we demonstrate the effectiveness of the proposed approach by conducting experiments on three face pose benchmark datasets: DPOSE multi-view dataset \cite{Subramanian2012An}, Head Pose Image Database (HPID) \cite{Gourier2004Estimating} and Biwi Kinect Head Pose Database (BKHPD)\cite{Fanelli2013Random}. Overview of them is shown in Table.

\begin{table}[!t]
\renewcommand{\arraystretch}{1.3}
\caption{An Example of a Table}
\label{table_example}
\centering
\begin{tabular}{|c||c||c||c|}
\hline
Name & Tasks & Training Set per Task & Testing Set per Task \\
\hline
DPOSE & 4 (Cameras) & 500 & 500\\
\hline
HPID & 3 (Views) & 465 & 465\\
\hline
BKHPD & 2 (Modals) & 400 & 465\\
\hline
\end{tabular}
\end{table}

DPOSE contains sequences acquired from 16 subjects, where the subject is either (i) rotating in-place at the room center, or (ii) moving around freely in a room, and moving their head in all possible directions. The dataset consists of over 50000 images, recorded from four cameras with overlapping field of view. Images have resolution $1024 \times 768$ and are stored in jpeg format. Head pan, tilt and roll measurements for various poses are recorded using an accelerometer, gyro, magnetometer platform strapped onto the head using an elastic band running down from the back of the head to the chin. We use the first subject, randomly choose 500 frames from each camera as the training set and the rest as the testing set. In this way, videos from each camera are treated as separate tasks. 

The Head Pose Image Database is a benchmark of 2790 monocular face images of 15 people with variations of pan and tilt angles from -90 to +90 degrees. For every person, 2 series of 93 images (93 different poses) are available. The purpose of having 2 series per person is to be able to train and test algorithms on known and unknown faces (cf. sections 2 and 3). People in the database wear glasses or not and have various skin color. Background is willingly neutral and uncluttered in order to focus on face operations. In our experiments, we use the first series of all the people as the training set and the second series as the testing set. The 93 poses of the first series are randomly divided into three views. 

In Biwi Kinect Head Pose Database, over 15K images of 20 people (6 females and 14 males - 4 people were recorded twice) are contained. For each frame, a depth image, the corresponding RGB image (both $640 \times 480$ pixels), and the annotation is provided. The face pose range covers about +-75 degrees yaw and +-60 degrees pitch. Ground truth is provided in the form of the 3D location of the head and its rotation. Because cheap consumer devices (e.g., Kinect) acquire row- resolution, noisy depth data, the authors recorded several people sitting in front of a Kinect (at about one meter distance). The subjects were asked to freely turn their head around, trying to span all possible yaw/pitch angles they could perform. Each frame is annotated with the center of the head in 3D and the head rotation angles. In our experiments, RGB images and depth images are treated as two modals. We randomly choose 20 degrees as the training set and the rest as the testing set.

To evaluate the performance of different methods, we estimate head pan and compute the differences between the estimation and the ground truth. Experiments are repeated 20 times to get the average performance and standard deviation. Besides, we make use of SeetaFace \cite{Kan2016Funnel} to locate the position of faces in DPOSE. To solve Eq. (18), we set $\lambda = 0.3$ and $\mu=0.5$.

\subsection{Comparison of different activation functions}

\begin{figure*}[!t]
\centering
\includegraphics[width=0.98\linewidth]{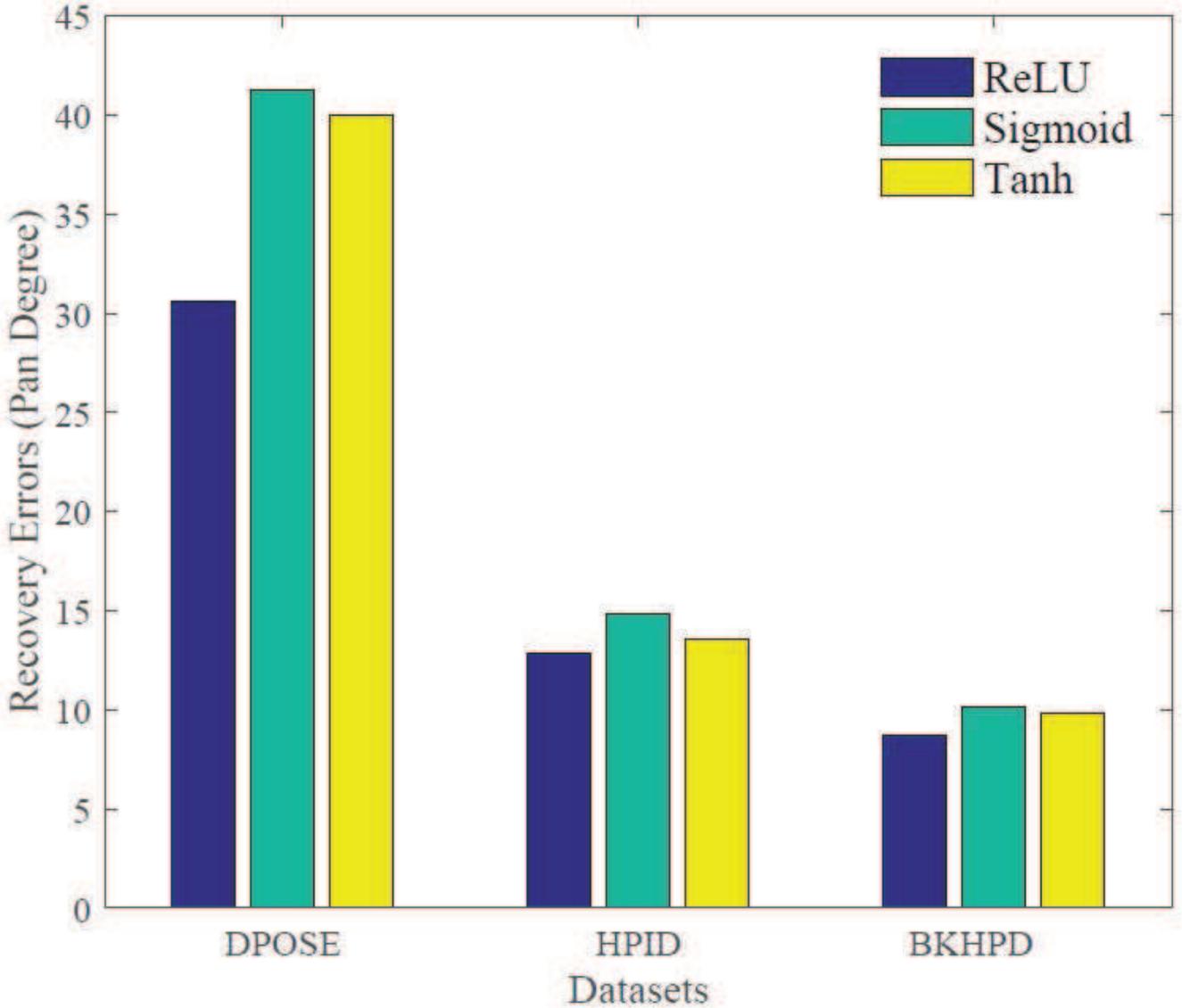}
\caption{Comparison of different activation functions. ReLU achieves the best performance.}
\label{activation}
\end{figure*}

As notated in III.C. there are several activation functions. They define the mapped output of a node in different ways. In this way, different activation functions may influence the performance. Therefore, we have tried ReLU, Sigmoid and Tanh. The results of three datasets are shown in Fig. \ref{activation}. We can figure out that ReLU achieves the best performance on all the datasets, which matches recent publications.

\subsection{Comparison of different multi-task loss functions}

Multi-task loss functions may also influence the performance of face pose estimation. Until now, a number of multi-task loss functions have been proposed to define the relationship among tasks. In our experiments, we compare the following loss functions:

\begin{itemize}
  \item Trace-Norm Regularized Learning with Least Squares Loss (LeastTrace) \cite{Ji2009An}: The loss function is defined as:
\begin{equation}
\begin{aligned}
&\arg \min_W \sum_{i=1}^t (0.5*norm(Y_i-X_i'*W(:,i))^2)+\\
&\rho_1 \parallel W \parallel_*,
\end{aligned}
\end{equation}
\noindent where $\parallel W \parallel_*=\sum(SVD(W,0))$ is the trace norm.

  \item L21 Joint Feature Learning with Least Squares Loss (LeastL21) \cite{Sch2007Multi}: The loss function is defined as:
\begin{equation}
\begin{aligned}
&\arg \min_W \sum_{i=1}^t (0.5*norm(Y_i-X_i'*W(:,i))^2)+\\
&opts.\rho_{L2}*\parallel W \parallel ^2_2+\rho_1 \parallel W \parallel_{2,1}.
\end{aligned}
\end{equation}

  \item Sparse Structure-Regularized Learning with Least Squares Loss (LeastLasso) \cite{Tibshirani2011Regression}: The loss function is defined as:

\begin{equation}
\begin{aligned}
&\arg \min_W \sum_{i=1}^t (0.5*norm(Y_i-X_i'*W(:,i))^2)+\\
&opts.\rho_{L2}*\parallel W \parallel ^2_F+\rho_1 \parallel W \parallel _1.
\end{aligned}
\end{equation}

  \item Incoherent Sparse and Low-Rank Learning with Least Squares Loss (Least- SparseTrace) \cite{Chen2012Learning}: The loss function is defined as:

\begin{equation}
\begin{aligned}
&\arg \min_W \sum_{i=1}^t (0.5*norm(Y_i-X_i'*W(:,i))^2)+\\
&\gamma*\parallel P \parallel_1, \\
&\mbox{subject to: } W=P+Q,\parallel Q \parallel_* \le \tau .&
\end{aligned}
\end{equation}

\noindent where $\parallel Q \parallel_* = \sum(SVD(Q,0))$ is the trace norm.

\end{itemize}

\begin{figure*}[!t]
\centering
\includegraphics[width=0.98\linewidth]{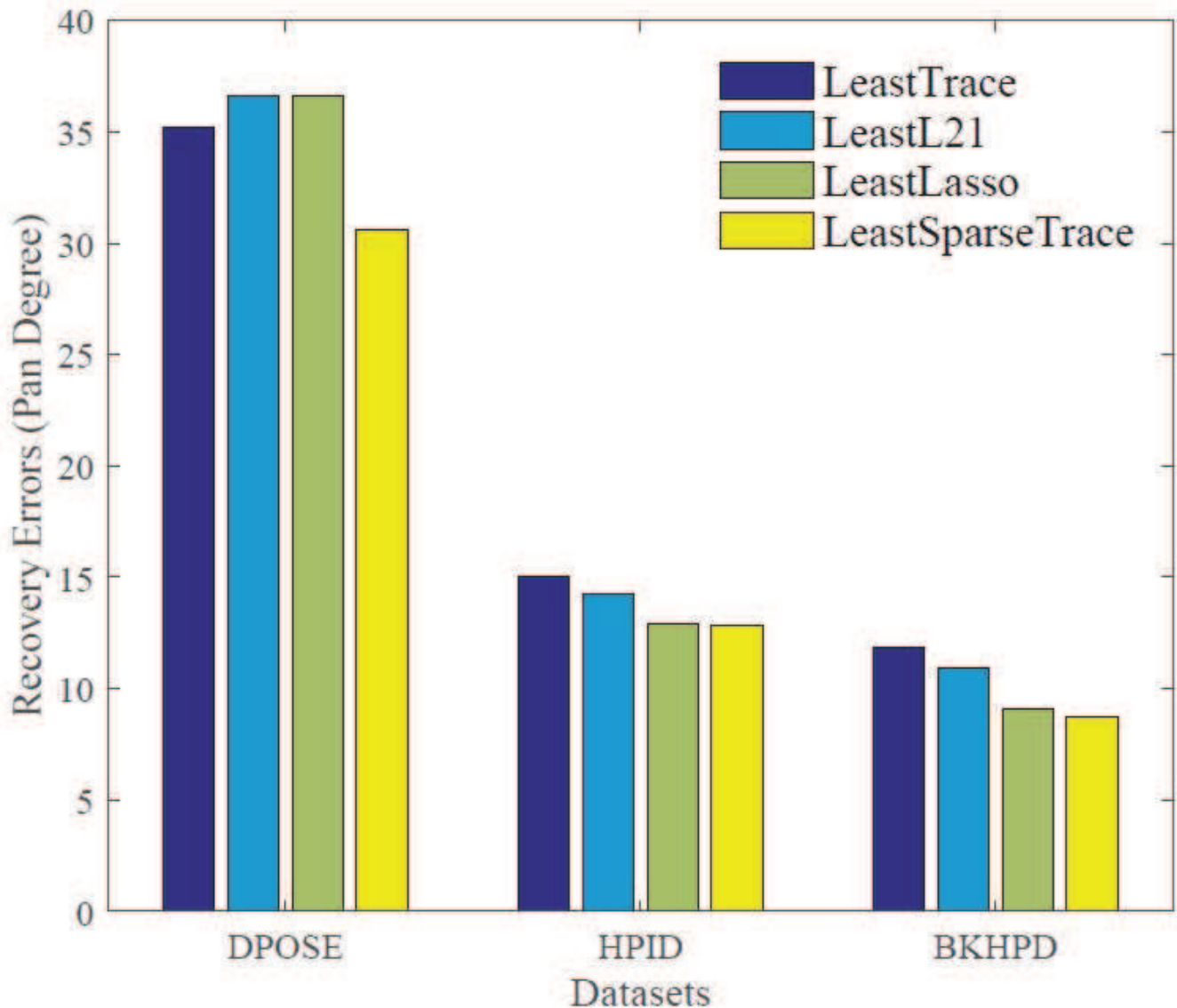}
\caption{Comparison of different loss functions of multi-task learning. LeastSparseTrace achieves the best performance.}
\label{loss}
\end{figure*}

We make use of MALSAR to conduct the performance \cite{zhou2011malsar}. The results are shown in Fig. \ref{loss}. We can clearly figure out that LeastSparseTrace outperforms the other loss functions. Due to the sparse constraints, LeastSparseTrace can improve the descriptive ability with features from different tasks. In our experiments, we use LeastSparseTrace as the loss function.

\subsection{Effectiveness of applying manifold learning and multi-task learning}

\begin{figure*}[!t]
\centering
\includegraphics[width=0.98\linewidth]{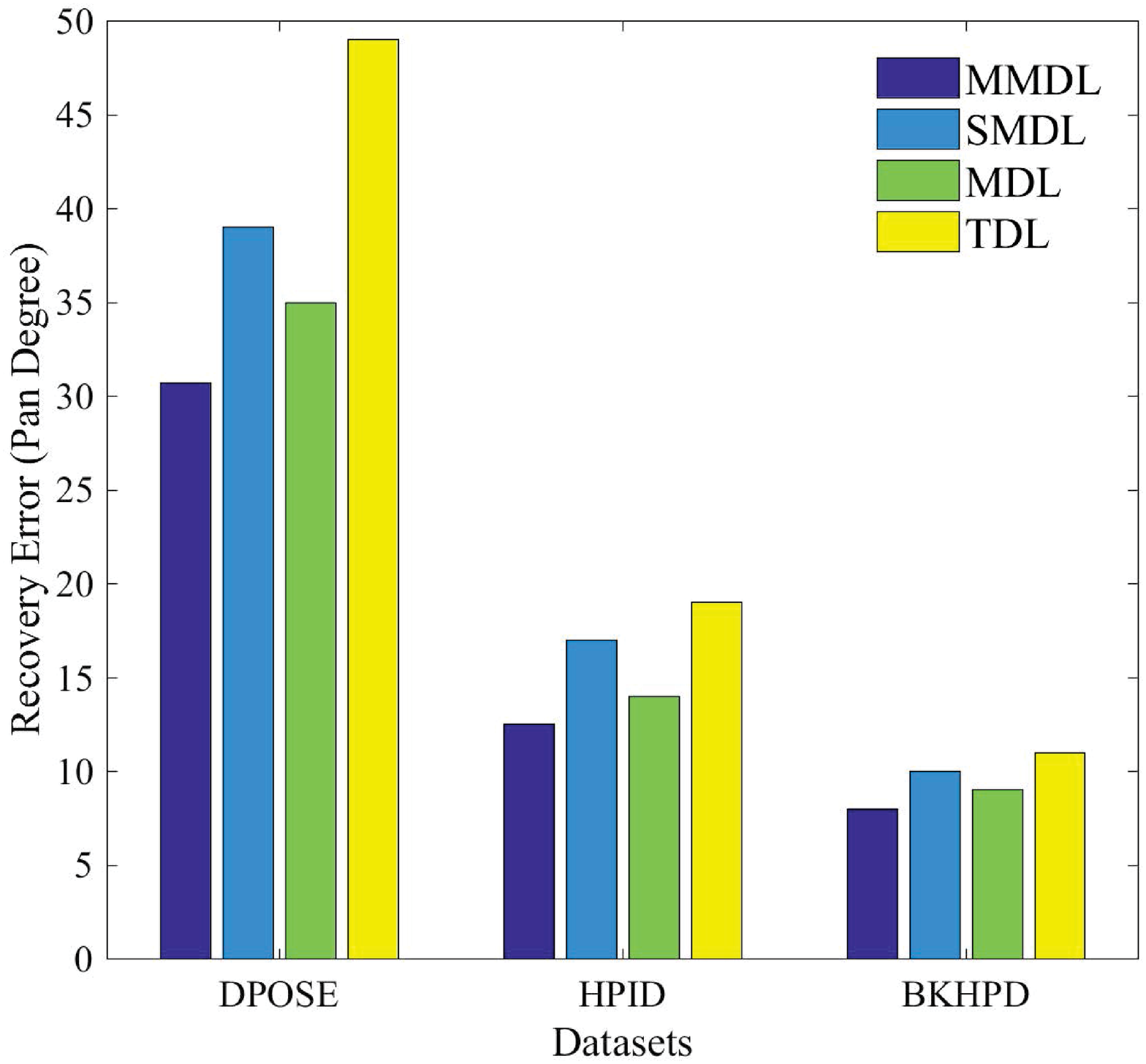}
\caption{Performance with and without manifold learning or multi-task learning}
\label{self}
\end{figure*}

In this part, we show the performance without manifold learning or multi-task learning to emphasize the effectiveness of the proposed $M^2DL$. Four different configurations are used. They are Multi-task Manifold Deep Learning $M^2DL$, Single-task Manifold Deep Learning (without multi-task learning)$SMDL$, Multi-task Deep Learning (without manifold learning) $MDL$ and Traditional Deep Learning (without manifold learning and multi-task learning) $TDL$. The performance is shown in Fig. \ref{self}. It can be seen that $M^2DL$ achieves the best performance, which indicates the effectiveness of the proposed method with manifold learning and multi-task learning.

\subsection{Comparison with state-of-the-arts}

For face pose estimation, we compare the following methods including the proposed multi-task manifold deep learning:

\begin{itemize}
  \item Multi-task Manifold Deep Learning  ($M^2DL$). The propose  method using multi-task learning and manifold regularized convolutional layer. ReLU is used as the activation function in the convolutional layer and LeastSparseTrace is used as the loss function in the multi-task learning.
  \item Linear regression (LR) \cite{Agarwal2006Recovering}: This recovers poses by direct linear regression against shape descriptor vectors extracted automatically from image silhouettes. Ridge regression and relevance vector machine (RVM) regression are applied; however, comprehensive experiments show that their performance is quite similar. In our comparison, RVM is adopted.
  \item Twin Gaussian Processes (TGP) \cite{Bo2010Twin}: TGP adopts Gaussian process priors on both covariates and responses, and estimates outputs by minimizing the Kullback-Leibler divergence between two Gaussian processes, which are modeled as normal distributions over finite index sets of training and testing examples.
  \item Salient Facial Structures (SFS) \cite{Gourier2004Estimating}. This is the baseline provided by HPID. In this method, the imagette containing the face is normalized in scale and orientation using moments provided by a face tracker. Each pixel in the face image is associated with an appearance cluster. One particular cluster stands for salient robust face structures which are: eyes, nose, mouth, chin. The authors have tried to extract and exploit a maximum of information provided by a single image of a face and to limit the loss of generality.
  \item Transfer Learning for Head Pose Classification (TLHPC) \cite{rajagopal2014exploring}. In this paper, the authors propose transfer learning solutions to overcome the adverse impact of changing attributes between the source and target data on face pose classification performance.
  \item Probabilistic High-Dimensional Regression (PHDR) \cite{Drouard2015Head}. This method maps HOG-based descriptors, extracted from face bounding boxes, to corresponding face poses. To account for errors in the observed bounding-box position, the authors learn regression parameters such that a HOG descriptor is mapped onto the union of a face pose and an offset, such that the latter optimally shifts the bounding box towards the actual position of the face in the image.
  \item Random Regression Forests (RRF) \cite{Fanelli2011Real}. The authors address the problem of face pose estimation from depth data, which can be captured using the ever more affordable 3D sensing technologies available today. To achieve robustness, pose estimation is formulated as a regression problem. While detecting specific face parts like the nose is sensitive to occlusions, learning the regression on rather generic surface patches requires enormous amount of training data in order to achieve accurate estimates.
  \item Head pose estimation using Perspective-N-Point (PnP). In this method, PnP is used to get the 6DOF pose of the head from point-correspondences. The correspondences are manually picked up beforehand. The 2D positions of Left Eye, Right Eye, Left Ear, Right Ear, Left Mouth, Right Mouth and Nose are used. Then 3D object¡¯s orientation is computed by solving a PnP (Perspective- N-Point) problem.
\end{itemize}

\begin{figure*}[!t]
\centering
\includegraphics[width=0.98\linewidth]{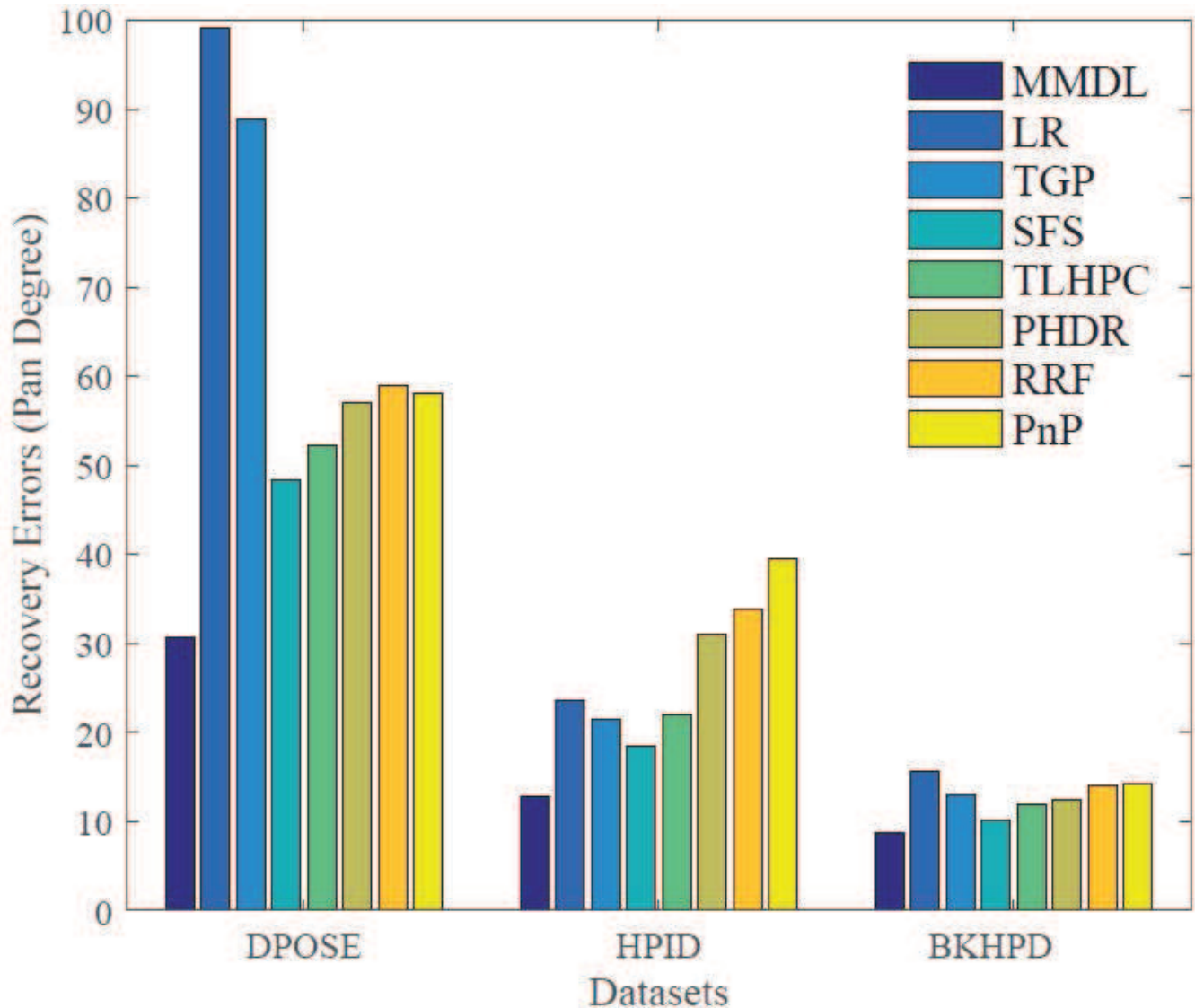}
\caption{Comparison with existing methods. The performance of the propose $M^2DL$ is better than state-of-the-arts.}
\label{state}
\end{figure*}

The results are shown in Fig. \ref{state}. Based on it, we can make the following summarizations:

\begin{enumerate}
  \item The performance of some general mapping learning methods such as LR and TGP cannot achieve satisfactory performance.
  \item The methods that look into the structure of faces such as SFS and RRF achieves better performance than the general methods.
  \item The performance of the propose $M^2DL$ is better than state-of-the-arts.
\end{enumerate}

\section{Conclusion}

In this paper, we propose a novel human face pose estimation method. It improves previous methods by employing deep learning and multi-task learning. First, the Manifold Regularized Convolutional Layers adopts manifold learning to compute the hidden relationship for neurons, with which we term better intrinsic representations. Second, we handle multi-modal features with multi-task learning. Each task handles a type of feature or a view. In this way, the mapping relationship can be simultaneously learnt. Finally, we can obtain an end-to-end mapping function from face images and poses. Experimental results on the datasets of DPOSE, HPID and BKHPD show that the proposed method outperforms previous methods of face pose estimation.

\ifCLASSOPTIONcompsoc
  \section*{Acknowledgments}
\else
  \section*{Acknowledgment}
\fi

This work was supported in part by the National Natural Science Foundation of China under Grants 61622205 and 61472110, the Zhejiang Provincial Natural Science Foundation of China under Grant LR15F020002, Australian Research Council Projects FT-130101457, DP-140102164, LP-150100671, the Fujian Provincial Natural Science Foundation of China (2016J01327, 2016J01324), the Fujian Provincial High School Natural Science Foundation of China(JZ160472) and the Foundation of Fujian Educational Committee (JAT160357).

\ifCLASSOPTIONcaptionsoff
  \newpage
\fi



\bibliographystyle{IEEEtran}
\bibliography{mda}

\end{document}